 \documentclass[final,5p,times,twocolumn]{elsarticle}

\usepackage{prletters}
\usepackage[T1]{fontenc} 

\usepackage{amssymb}
\usepackage{url} 

\journal{Pattern Recognition Letters}

\begin{document}

\begin{frontmatter}



\title{mEBAL2 Database and Benchmark: Image-based Multispectral Eyeblink Detection}


\author[inst1]{Roberto Daza}
\ead{roberto.daza@uam.es}
\author[inst1]{Aythami Morales}
\ead{aythami.morales@uam.es}
\author[inst1]{Julian Fierrez}
\ead{julian.fierrez@uam.es}
\author[inst1]{Ruben Tolosana}
\ead{ruben.tolosana@uam.es}
\author[inst1]{Ruben Vera-Rodriguez}
\ead{ruben.vera@uam.es}

\affiliation[inst1]{organization={Biometrics and Data Pattern Analytics Laboratory, Universidad Autonoma de Madrid},
            addressline={ Calle Francisco Tomas y Valiente, 11, Campus de
Cantoblanco}, 
            city={Madrid},
            postcode={28049}, 
            country={Spain}}

\begin{abstract}
This work introduces a new multispectral database and framework to train and evaluate eyeblink
detection in RGB and Near-Infrared (NIR). Our contributed dataset (mEBAL2, multimodal EyeBlink and Attention Level estimation, Version 2) is the largest existing eyeblink database, representing a great opportunity to improve data-driven multispectral approaches for blink detection and related applications (e.g., attention level estimation). mEBAL2 includes 21,100 image sequences from 180 different students (more than 2 million labeled images in total) while conducting a number of e-learning tasks of varying difficulty or taking a real course on HTML initiation through the edX MOOC platform. mEBAL2 uses multiple sensors, including two Near-Infrared (NIR) and one RGB camera to capture facial gestures during the execution of the tasks, as well as an Electroencephalogram (EEG) band to get the cognitive activity of the user and blinking events. Furthermore, this work proposes 3 data-driven approaches as benchmarks for blink detection on mEBAL2, where the architecture based on Convolutional Long Short-Term Memory (ConvLSTM) achieved performances of up to 99\%.  The experiments explored whether combining RGB and NIR spectrum data improves blink detection in training and architectures that merge both types of data. Experiments showed that the NIR spectrum enhances results, even when only RGB images are available during inference. Finally, the generalization capacity of the proposed eyeblink detectors, along with state-of-the-art eyeblink detection implementations, is validated in wilder and more challenging environments like the HUST-LEBW dataset to show the usefulness of mEBAL2 to train a new generation of data-driven approaches for eyeblink detection.
\end{abstract}


\begin{highlights}
\item We are presenting mEBAL2, the largest eyeblink database.
\item Eyeblink detection enhanced through the combination of NIR and RGB images.
\item Competitive eyeblink detection, up to $99\%$ accuracy in e-learning.
\item Validation in challenging wild environments, achieving state-of-the-art results.

\end{highlights}


\end{frontmatter}


\section{Introduction}
\label{sec:Introduction}
The act of involuntary closing and opening of the eyelids periodically is
defined as eyeblink. The eye is one of the most important organs in the human
facial structure for image processing applications from behavior analysis to
biometric
identification~\cite{2019_HBookSelfie_SuperSelfieFaceIris_Alonso,alonsofernandez2023periocular}.
The eyeblink has proven to be a valuable indicator in various fields such as
ocular activity, attention, fatigue, emotions, etc., for this reason, eyeblink
detection based on image processing has become essential regarding applications
involving human behavior analysis, such as driver fatigue
detection~\cite{bergasa2006real}, attention level
estimation~\cite{daza2021alebk,daza2023matt}, dry eye syndrome
recovery~\cite{rosenfield2011computer}, DeepFakes
detection~\cite{jung2020deepvision}, among others.

In the e-learning field, eyeblink detection can be a valuable tool to address
certain limitations, particularly, when combined with the latest e-learning
platforms~\cite{hernandez2019edbb,daza2023edbb,becerra2023m2lads} that allow
collecting students' information to improve security, to guarantee a safe and
personalized evaluation, and to adapt dynamically the contents and
methodologies to different needs. Eyeblink detection is a useful tool to
improve e-learning platforms and get high-quality online education, for at
least two reasons. First, since the $70$s, studies relate the eyeblink
rate with cognitive activity like
attention~\cite{bagley1979effect,holland1972blinking}. Recent research suggests
that lower eyeblink rates can be associated with high attention periods while
higher eyeblink rates are related to low attention
levels~\cite{daza2021alebk,daza2023matt,daza2020mebal}.
And second, blink detection can be used in the detection of fraud/cheating/lies and combined with other features like heart rate, gaze tracking, micro-gestures or blood oxygen saturation, can improve the trustworthiness of e-learning platforms.

\begin{figure}[!t]
 \centering
  \includegraphics[width=\linewidth]{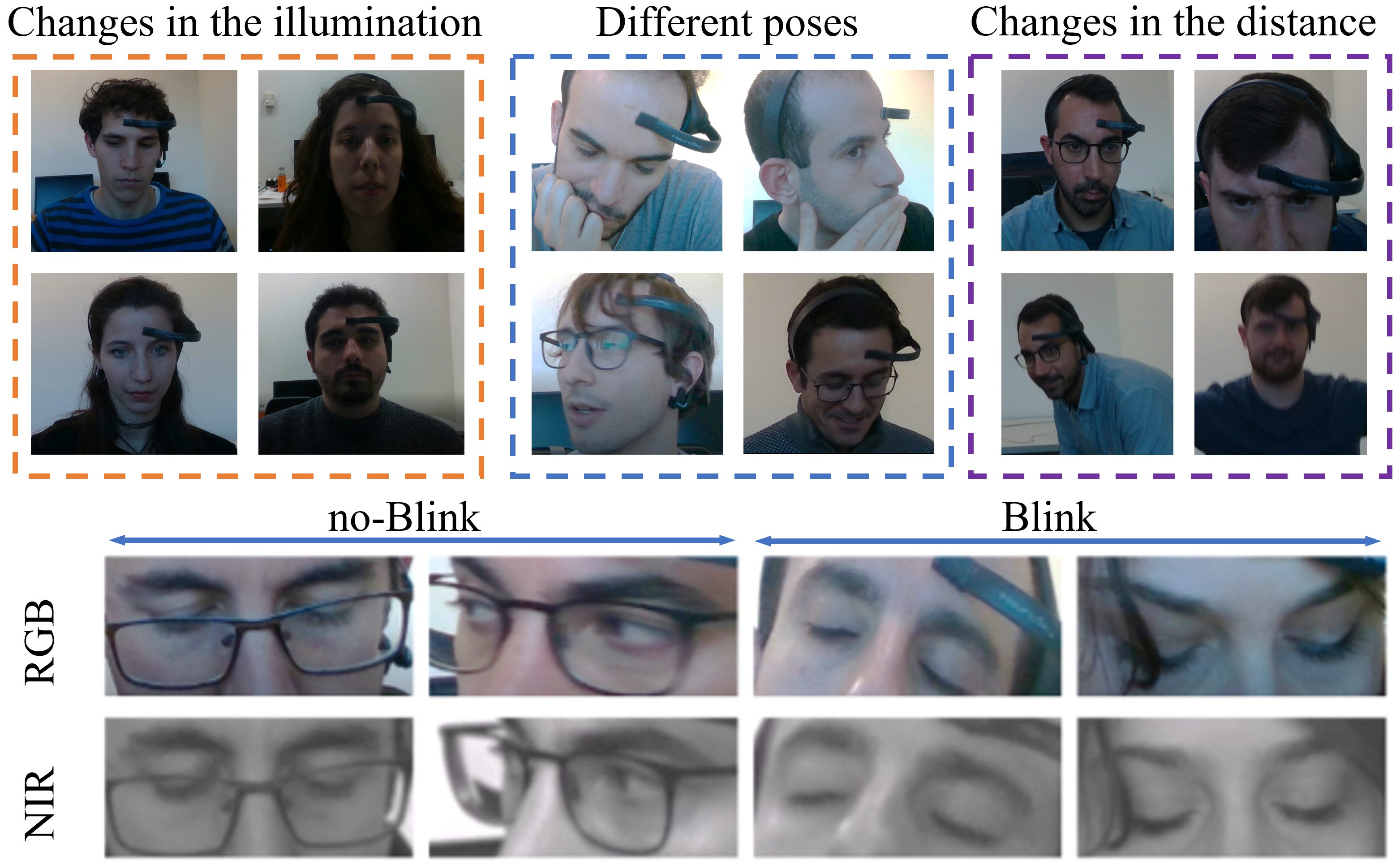} 
  \caption{Different examples from mEBAL2. (top) Sequence images with variations in illumination, posing, and distance to the camera. (bottom) Examples of eyeblink and no-blink with RGB and NIR images.}
   \label{fig:mEBAL V2}
\end{figure}

Over the past few years, significant progress has been made in eyeblink
detection~\cite{daza2021alebk,daza2020mebal,hu2019towards,zeng2023real} thanks
to the rise of new computer vision technologies and deep learning techniques,
facilitating improved detection of the region of interest and subtle movements
of the eyelids, even in challenging conditions such as low lighting and
variable poses.

However, the state of the art demonstrates that public eyeblink detectors based
on image processing are far from resolving the eyeblink detection problem,
HUST-LEBW~\cite{hu2019towards} and MPEblink~\cite{zeng2023real} are recent
surveys that demonstrate the need for research in this area. At the moment,
there are very few public data-driven algorithms (e.g.,~neural networks),
mainly due to the lack of large eyeblink databases. Most existing datasets
useful for research in this area have only a few hundred samples, representing
a strong restriction to training data-driven approaches (e.g.,~deep learning).
Also, current public databases restrict their samples to RGB cameras, without
using other sensors that have proven to be useful in similar tasks such as NIR
cameras in gaze tracking or iris and pupil
detection~\cite{2022_INFFUS_Periocular_Alonso}.

Considering all of the above, this work aims to provide resources to train and evaluate eyeblink detectors, investigate the potential utility of the NIR spectrum for eyeblink detection, and evaluate new data-driven approaches in realistic environments. The main contributions are:

\begin{itemize}
\item We present a new version of the mEBAL database called  mEBAL2\footnote{\url{https://github.com/BiDAlab/mEBAL2}}: a multimodal database for eyeblink detection and attention level estimation obtained from an e-learning environment. This database is the largest existing public eyeblink database. This database includes videos from $180$ students, with 21,100 labeled image sequences (10,550 eyeblinks and 10,550 no-blink events), more than 2.4 million labeled frames, and students' cognitive activity labels synchronized with all the eyeblink data. Additionally, the new mEBAL2 contains variations on illuminations, poses, distances between user and camera, objects over the face (glasses, hair, hand occlusion, etc.), physical activity, and other naturally-occurring factors.  Fig.~\ref{fig:mEBAL V2} shows some examples from mEBAL2.

\item Specific tasks were designed to provoke changes in the students' cognitive activity like mental load, attention, visual attention, etc. The experiments comprise data from two groups divided into 60 and $120$ users. The second group attended a Massive Open Online Course (MOOC) offered by the Universidad Autonoma de Madrid (UAM). This approach provides a real-world environment.   

\item Our architecture, presented in mEBAL~\cite{daza2020mebal}, was trained on
 the proposed mEBAL2 dataset using both RGB and NIR frames.  The results
  demonstrate that NIR images improve the eyeblink detector's
  training process and outperform the results even when only RGB
  images are available during the inference.

\item Two new data-driven approaches are proposed: An eyeblink detection at frame level based on CNNs that combines all spectral information, and an eyeblink detection at video-sequence level based on ConvLSTM. Additionally, two benchmarks are introduced within mEBAL2: (1) for blink detection at video-sequence level, and (2) for blink detection at frame level.

\item Finally, experiments were conducted to showcase the ability of mEBAL2 to train data-driven approaches that can generalize to unseen scenarios. The public benchmark HUST-LEBW, captured in uncontrolled environments, was utilized for evaluation. The results highlight mEBAL2's proficiency in developing new eyeblink detectors that effectively adapt to previously unseen scenarios.
\end{itemize}

A preliminary version of this article was presented in~\cite{daza2020mebal}.
This article significantly improves~\cite{daza2020mebal} in various aspects: 

\begin{itemize}

\item An improved version of the mEBAL database with $180$ users and
 7,550 additional eyeblink events, which now includes 5 times more users
  and 3.5 times more eyeblink events compared with
  ~\cite{daza2020mebal}. Furthermore, a new real e-learning
  environment has been added. 

\item For the first time, we trained our architecture proposed in mEBAL on mEBAL2, using RGB, NIR, and both images. We conducted a new study to verify whether using NIR images during training could enhance eyeblink detection.

\item Two new blink detectors are proposed, being the most accurate systems on mEBAL2, surpassing the previous version presented in mEBAL.

\item Evaluations were conducted on the public HUST-LEBW database \cite{hu2019towards} using the same eyeblink detector
  introduced in mEBAL, but with training on the new version. This
  resulted in an error reduction of 37.31\% for the left eye and
  27.85\% for the right eye.

\end{itemize}

The rest of the paper is organized as follows. Section $2$ summarizes works related to eyeblink detection. Section $3$ presents the mEBAL2 database. Section $4$ describes the
architecture proposal for the eye’s localization and the architectures for the eyeblink verification. Section $5$ describes the experiments. Finally, Section $6$ provides conclusions and future work.

\section{Related Works} \label{sec:relatedWorks}

\subsection{Eyeblink detection databases}

\begin{table}[t]
\centering
\setlength\tabcolsep{7pt}
\renewcommand{\arraystretch}{1.1}
\caption{State of the art of eyeblink database.}
  \label{tab:datasets}

  \begin{tabular}{|c|c|c|c|c|c|c|c|}
    \hline
    \bf{DB.} & \bf{Year} & \bf{Blinks}  & \bf{Users}  & \bf{Att.} & \bf{Sensors} \\
    \hline \hline
    \cite{Talkingface}& NA &  $61$ & $1$  & No & RGB \\
    \hline
    \cite{pan2007eyeblink} & 2007 & $255$ & $20$ & No  & RGB \\
    \hline
    \cite{drutarovsky2014eye} & 2014 & $353$ & $4$  & No  & RGB \\
    \hline
    \cite{radlak2015silesian} & 2015 & $300$ & $5$   & No  & RGB \\
    \hline
    \cite{hu2019towards} & 2019 & $381$ & $172$   & No  & RGB\\ 
    \hline
     \cite{daza2020mebal} & 2020& 3000 & $38$  & Yes  & 1 RGB - 2 NIR \\
    \hline
    \cite{zeng2023real} & 2023& 8748 & NA  & No  & RGB\\
    \hline
   \textbf{Ours} & \textbf{2024} & \textbf{10550} & \textbf{180} & \textbf{Yes} & \textbf{1 RGB - 2 NIR}\\
    \hline
  \end{tabular}
\end{table}

There are several well-known databases for blink detection, such as Talking
Face~\cite{Talkingface}, ZJU~\cite{pan2007eyeblink},
Eyeblink8~\cite{drutarovsky2014eye}, and Silesian5~\cite{radlak2015silesian},
which all share a common limitation: the small number of users and blinks (see Table~\ref{tab:datasets}).  This limited amount of data implies a significant
restriction for data-driven approaches during training. The reported results
from published evaluations  show saturated scores close to $99$\% of
accuracy due to the reduced number of eyeblink
samples~\cite{Talkingface,pan2007eyeblink,drutarovsky2014eye,radlak2015silesian}.

The previous databases are characterized by acquisition setups under controlled
environments. The HUST-LEBW database~\cite{hu2019towards} was proposed to
explore the detection of eyeblinks in unconstrained scenarios. This database
uses eyeblink video clips from $20$ commercial movies (The Matrix, Lord
of the Rings, etc.). It contains indoor and outdoor examples with
scene/illumination changes and varying human poses, similar to a wild
environment. HUST-LEBW consists of $172$ actors with $381$
eyeblinks. It is divided into training with $254$ eyeblinks ($253$
right eye and $243$ left eye) and testing with $127$ eyeblinks
($126$ right eye and $122$ left eye). The database is unbalanced
because the training includes $190$ no-eyeblink ($190$ right eye
and $181$ left eye) and testing includes $98$ no-eyeblink
($98$ right eye and $98$ left eye). The major drawback is the
small training set for data-driven approaches ($448$ samples). A small
training set may cause a poor generalization because a few eyeblink examples
are not able to handle different head poses, illumination, hair on the eyes,
makeup (overstated use of makeup), etc.

The MPEblink database~\cite{zeng2023real} was published recently, and also
collected eyeblink data from a wild environment, similar to the
HUST-LEBW~\cite{hu2019towards}. This database contains $686$ short video
clips (7.1--85.9 s) of $86$ different movies. 8,748 eyeblink events were
labeled in total, and each video has different eyeblink events from different
people.

mEBAL~\cite{daza2020mebal}  is a previous version of mEBAL2,  also captured in
an e-learning environment in a multispectral setup (RGB and NIR cameras). mEBAL
has $38$ users with 3,000 eyeblinks. mEBAL2 comprises three times more
users and blink samples with more than 2.4 million frames recorded from
$180$ MOOC users.
Table~\ref{tab:datasets} summarizes the main differences between mEBAL2 and the databases used in the literature.

\subsection{Eyeblink detection methods} \label{sec:methods}

An eyeblink is a sequence of ``eyes open - eyes closed - eyes open'' that occurs in a short time. For this reason, eyeblink detectors can be categorized into two groups: 
\begin{itemize}

\item Eyeblink detection at frame level (image-based eyeblink detection):  In
 this case, the methods classify each frame in open, closed, or the
  degree of eye closure. Then a sequence of events is defined to
  detect the eyeblink action like ``open/closed/open'',
  ``open/partially-closed/closed/open'', etc. Some methods
  based on CNN have been proposed for this group. For example,
  in~\cite{anas2017online}, Anas et al presented two methods
  based on CNN. The first method had two states (open or closed
  eyes), and the second had a third state for a partially opened
  eye. The authors evaluated ZJU and Talking Face datasets
  getting 93.72\% and $100$\% blink detection accuracy
  using F1 scores. Phuong et al~\cite{phuong2022eye}  presented a
  model based on the Eye Aspect Ratio (EAR), innovating with a custom EAR threshold.  

\item Eyeblink detection at video-sequence level: This approach uses in a
 holistic manner the information obtained from a sequence of frames,
  which is classified entirely as blink or no-blink. Some
  researchers like Soukupov\`a et al~\cite{soukupova2016eye} used
  $13$ consecutive frames as input to extract the EAR
  using facial landmarks. $13$ EARs were concatenated as
  input to an SVM to classify between blink or no-blink.
In addition, Huge et al~\cite{hu2019towards} proposed a model based on a
  Multi-Scale LSTM (MS-LSTM). Appearance and motion features were
  extracted in each frame sequence to classify eyeblinks. The
  MS-LSTM model outperformed some state-of-the-art algorithms in
  wild environments.
Zeng et al~\cite{zeng2023real} presented the InstBlink model recently, which
  used Query-based methods~\cite{carion2020end} as its foundation
  to obtain the face bounding box and eyeblink labels, without
  using the eye localization method. 
Zeng et al~\cite{zeng2023eyelid}, in a recent work, proposed an approach that
  captured eyelid movement features to differentiate between
  blink and no-blink. The network has an
  architecture with an attention module to
  generate an attention map, which is fed into a CNN model to
  jointly learn the appearance and movement features in each
  frame. Finally, the features extracted are
  used as input to an LSTM model.

\end{itemize}

\section{mEBAL2 Database} \label{sec:database}

mEBAL2 contains synchronized information from multiple sensors while the
students use an interface designed for e-learning tasks. mEBAL2 acquisition is
based on the works of Hernandez  et al in~\cite{hernandez2019edbb} and Daza
 et al in~\cite{daza2023edbb}. The authors proposed a platform for remote
education assessment called edBB (Biometrics and Behavior for Education).
A multimodal acquisition framework was designed to monitor cognitive and eyeblink activity during e-learning tasks. 
mEBAL2 includes 21,100 events (10,500  blinks and 10,500  no-blinks) from 180 students/sessions. The session duration varies from 15 to 40~min. Each eyeblink event has 19 frames using three cameras: one RGB and two NIR cameras. This database contains 2,405,400 frames (3 cameras $\times$ 19 frames $\times$ 21,100 events $\times$ 2 eyes), making it the largest existing eyeblink database. 

Therefore, mEBAL2 provides a dataset consisting of $540$ long-duration videos ($1$ RGB video and $2$ NIR videos per session). Each video comes along with the facial bounding box information, $68$ facial landmarks, and cropped eye regions for each frame. Furthermore, the dataset includes timestamps for eyeblink and no-eyeblink events and a total of 21,100 cropped samples. Additionally, the dataset provides EEG band information, including attention level, meditation level, $5$ electroencephalographic channels, and eyeblink intensity measures. Finally, mEBAL2 contributes two subsets: (\textit{i})  For frame-level blink detectors, a subset based on the eye state, with 21,000 frames for open eyes and 21,000 frames for closed eyes, is provided. This offers a resource where frame-level blink detectors can be trained and evaluated. (\textit{ii})  For video-sequence level blink detectors, the subset contains 10,500 blinks and 10,500 no-blinks.
 
 The acquisition setup uses the following sensors (see~\cite{daza2023edbb} for
a graphical representation of the setup):

\begin{itemize}

 \item An Intel RealSense (model D435i), which contains $1$ RGB and
  $2$ NIR cameras.   The $3$ cameras are configured to
  $30$ Hz and $1280\times720$ resolution. According to the
  Harvard Database of Useful Biological
  Numbers~\cite{schiffman1990sensation}, an average eyeblink
  ranges between $100$ ms--$400~{\rm ms}$.   Our experience with
  mEBAL2 reveals that normally eyeblink duration is between
  $198$--$263~{\rm ms}$.   Therefore, bearing in mind previous
  studies, our experience, and the setup settings, the average
  eyeblink can last between $3$ to $13$ frames.
 
 \item An EEG headset by NeuroSky, which measures the power spectrum density of
  $5$ electroencephalographic channels ($\alpha, \beta, \gamma, \delta, \theta$). EEG
  measures the voltage signals produced usually by synaptic
  excitations of the dendrites of pyramidal cells in the top
  layer of the brain cortex~\cite{kirschstein2009source}.  The
  signals are produced mainly by the number of neurons and fibers
  fired synchronously~\cite{hall2020guyton}. Eyeblinks introduce
  artifacts that can be easily recognized in EEG signals. In this
  dataset, the EEG band was used to generate the initial ground
  truth necessary to label the eyeblink events. 
 
\end{itemize}

 The eyeblinks were labeled using a semi-supervised approach.  
For that labeling, we first used the eyeblink information provided automatically by the EEG band as candidates for ground truth, and then a human manually checked all the detected events. Without human intervention, the number of eyeblinks detected was 21,484. After the human intervention, the eyeblinks were reduced to 12,032, where 1,482 were labeled as possible eyeblinks and the remaining eyeblinks were considered ground truth. Each eyeblink in mEBAL2 contains $19$ frames in total.

\subsection{Tasks}

The database was divided into two groups. The first group of 60 students did a series of tasks carefully designed to reach certain goals, and the second group of 120 students did a real lesson from a MOOC entitled ``Introduction to Development of Web Applications'' (WebApp), available in the edX platform. The lesson is about introduction to HTML coding, where students perform different tasks including watching videos, reading documents, reading and writing HTML code, and performing a final exam.

The tasks for both groups were designed with two goals. First, to generate
changes in the students' cognitive activity such as mental load, attention,
visual attention, etc., looking to cause variations of the eyeblink rate.
Second, to generate a realistic setting of online assessments. The tasks can be
categorized into five groups (see~\cite{daza2023edbb} for a video
demonstration\footnote{\url{https://www.youtube.com/watch?v=JbcL2N4YcDM}}):

\begin{itemize}
\item Enrollment form: Student's data are obtained here. This simple task targets a relaxed state with attention levels between normal and low.

\item Logical questions: These require more complex interactions, and some of them include crosswords and mathematical problems for the first group. For the MOOC course, some activities involve writing HTML codes and generating more efficient ones. 

\item Visual tasks: These demand visual attention from the students under different situations, such as watching pre-recorded classes, describing images, and detecting errors in HTML code.

 \item Reading tasks: Reading documents has proven to have an impact on eyeblink rates and it is highly common in e-learning environments.

\item Multiple choice questions: These are essential to help evaluate the
 students on assessment platforms and most Learning Management Systems
  provide templates to perform these
  assessments~\cite{govindasamy2001successful}.  

\end{itemize}

\section{Proposed Eyeblink Detector} \label{sec:Eye Blink Detector} 

We propose architectures for eyeblink detection: (1) ROI localization, which is commonly used in eyeblink detectors, and (2) Eyeblink verification.

\subsection{ROI detection}

A sequential approach using deep learning is proposed for ROI detection in 5
steps: (\textit{i}) Face detection: using the well-known RetinaFace
Detector~\cite{deng2020retinaface}, a robust single-stage face detector that
uses extra-supervised and self-supervised multi-task learning. (\textit{ii})
Landmark detection: using a $68$ SBR landmark
detector~\cite{dong2018supervision}, based on
VGG-$16$~\cite{simonyan2014very} and Convolutional Pose Machines
stages~\cite{wei2016convolutional}. (\textit{iii}) Face Alignment: The Dlib
library~\cite{dlib} is used to align both eyes parallel to a horizontal
line~\cite{2015_FSI_FacialSoftBio_Pedro}. The alignment is performed utilizing
five landmarks (two eyes, the nose, and two at the mouth). The inclination
angle formed by the eye landmarks is calculated in the face rotation process,
and an affine transformation is applied to correct the tilt. (\textit{iv}) Data
quality: Using detectors' probabilities ($p_{ROI}$), ROI quality can be
assessed. Two different probabilities are calculated, one before alignment
($p_{ROI_{pre}}$) and another after ($p_{ROI_{post}}$), leading to three potential
decisions: maintaining alignment ($0.60 \leq p_{ROI_{post}}$), not maintaining it ($0.60>p_{ROI_{post}}$
or $ \frac{2}{3} \times p_{ROI_{pre}}> p_{ROI_{post}}$), or discarding the frame ($0.25>p_{ROI_{post}}$ and $0.25>p_{ROI_{pre}}$).
Consequently, the process involves recalculating the eye landmarks. A
$p_{ROI}$ below 0.25 indicates a failed detection or a turned head.
(\textit{v}) Eye cropping: Finally, each eye was cropped using the ROI's
information from the landmark detectors. Later, all eyes were resized to
$50 \times 50$.

\subsection{Eyeblink verification architecture}

The mEBAL2 experimental framework includes $2$ blink detectors at
frame-level based on CNN architectures and $1$ blink detector at
video-sequence level based on a ConvLSTM
architecture~\cite{shi2015convolutional}. 

\textbf{One-Eye ConvNet architecture (OE-ConvNet)} \cite{daza2020mebal}: This
architecture is based on the popular VGG16 neural network
model~\cite{simonyan2014very} (see mEBAL~\cite{daza2020mebal} for details). The
architecture is formed by $3$ convolutional layers with ReLU activation
($32/32/64$ filters of size $3 \times 3$), with $3$ max pooling
layers between them. The last layer is used as input for a dense layer of
$64$ units with ReLU activations and 0.5 of dropout. In this work, we
have adapted~\cite{daza2020mebal} for different training scenarios including
RGB and NIR spectrums. During the training process, the RGB and NIR images were
introduced in the training batch depending on the scenario. 

\textbf{Left NIR}  $+$  \textbf{Right NIR}  $+$  \textbf{RGB ConvNet architecture (LI-RI-RGB-ConvNet)}: We propose a late fusion~\cite{2018_INFFUS_MCSreview1_Fierrez} using the 3 channels of the Intel RealSense information (1 RGB and 2 NIR). It consists of six inputs (2 eyes $\times$ 3 cameras), and each input layer has $50 \times 50 \times C$ dimensions, where $C$ is the number of channels for each used spectrum  ($3$ for RGB and $1$ for NIR). All $6$ inputs were connected to $6$ different convolutional blocks, with the same characteristics of the OE-ConvNet. The outputs of the $6$ convolutional blocks were concatenated and connected to a dense layer of $64$ units (ReLU activation) and an output layer with one unit (sigmoid activation). A dropout of 0.5 was used. 

\textbf{One-Eye ConvLSTM architecture (OE-ConvLSTM)}: This architecture processes sequences of $10$ input images. It consists of three ConvLSTM layers using recurrent activation with hard sigmoid, $32$ filters with a kernel size of $3\times3$ (tanh activation) followed by a batch normalization layer and a max pooling layer, excepting the third layer, which changes the number of filters from $32$ to $64$. Finally, the architecture incorporates a dense layer of $64$ units with ReLU activations. Classification between an eyeblink and no-blink is performed by a final output layer with one unit and sigmoid activation. In addition, dropout (0.5) is used.  

OE-ConvNet and  OE-ConvLSTM offer significant advantages as they were trained to model each eye separately. Therefore, they allow for detecting eyeblinks in side poses, even when one of the eyes is occluded. On the other hand, the LI-RI-RGB-ConvNet takes advantage of all the information provided by the RealSense camera.

\section{Experiments and Results} \label{sec:expResults}

\subsection{Experimental protocol}

The proposed architectures were trained using mEBAL2 from scratch with a batch size of $32$, an Adam optimizer (0.001 learning rate) and a binary cross-entropy loss.

The mEBAL2 benchmark includes a leave-one-out cross validation protocol with one user for testing and the remaining users for training. The process was repeated for each user in the database and the obtained results were averaged. The decision threshold was fixed to the point of Equal Error Rate (EER), in which the False Positive and False Negative rates in blink detection are equal.

The generalization ability of the eyeblink detector trained with mEBAL2 was
evaluated on the public benchmark HUST-LEBW~\cite{hu2019towards}. The HUST-LEBW
dataset comprises videos obtained from films characterized as in-the-wild
completely different to the mEBAL2 environment (e-learning).

\subsection{Experiments: mEBAL2 Benchmark}

\subsubsection{mEBAL2: Blink detection at frame-level}
Table~\ref{tab:Results} ppresents the mEBAL2 benchmark results of the OE-ConvNet under different training scenarios (e.g.,~RGB or NIR, different eyes). The results show detection accuracies up to 97.30\% in the RGB and 93.94\% in the NIR images. Results also show how training a specific detector for each eye leads to a slight improvement compared to training one for both eyes.

One of our goals was to understand if the NIR images could serve to improve current eyeblink detectors. For this reason, we trained OE-ConvNet using RGB, NIR, or both.  The results in 
Table~\ref{tab:Results} show that eyeblink detection with RGB images is more
accurate than the NIR images with a difference of 2.49\%  approximately. The
results with the left NIR camera are similar to the right NIR camera. We
trained our OE-ConvNet approach using data from all $3$ images
($1$ RGB  $+$  $2$ NIR) and evaluated the performance over the RGB
images, adopting a similar approach to~\cite{fu2021siames}. The batch of size
$32$ was generated with both RGB and NIR images. The NIR images (size
$50\times50\times1$) were expanded to size $50\times50\times3$, which became the input size for
our architecture. As we can see in 
Table~\ref{tab:Results}, there is an improvement of 0.41\% when all three images are used for training (FS  $=$  Full Spectrum), compared to the results obtained when training with RGB only. 


\begin{table}[t!]
\centering
\setlength\tabcolsep{7pt}
\renewcommand{\arraystretch}{1.1}
\caption{Comparison of OE-ConvNet on mEBAL2 under different training/evaluation scenarios. The Eyes column denotes the eyes used for training/evaluation. FS: Full Spectrum (RGB and both NIR cameras). NIR$_{x}$: x is the Camera (R: Right, L: Left, or B: Both).}

\label{tab:Results}
\begin{tabular}{|c|c|c|c|c|}
\cline{1-5}

\multicolumn{2}{|c|} {\textbf{Eyes}} &   \textbf{Training} &\textbf{Evaluation} & \textbf{Acc}  \\ \hline \hline

\multicolumn{2}{|c|}{\multirow {5}{*} {Both}} & RGB & RGB & $0.9615$\\ \cline{3-5}
\multicolumn{2}{|c|}{} &  NIR$_{R}$ & NIR$_{R}$ & $0.9373$              \\ \cline{3-5}
\multicolumn{2}{|c|}{ } &  NIR$_{L}$ & NIR$_{L}$ & $0.9360$  \\ \cline{3-5}
\multicolumn{2}{|c|}{} & NIR$_{B}$ & NIR$_{B}$ & $0.9394$   \\ \cline{3-5}

\multicolumn{2}{|c|}{\multirow {5}{*} {}} & FS (RGB+NIR$_B$) & RGB & $0.9656$  \\  \hline \hline

\multicolumn{2}{|c|}{Left} & RGB  & RGB & $0.9730$   \\ \hline
\multicolumn{2}{|c|}{Right} & RGB  & RGB & $0.9669$  \\ \hline 

\end{tabular}

\end{table}

\begin{table}[t!]
\centering
\setlength\tabcolsep{7pt}
\renewcommand{\arraystretch}{1.1}
\caption{Eyeblink detection accuracy at frame-level obtained by OE-ConvNet (FS), LI-RI-RGB-ConvNet, and two existing blink detectors \cite{phuong2022eye,soukupova2016eye}.}
\label{tab:Benchmark}
\begin{tabular}{|l|c|}
\hline
\textbf{Method} & \textbf{Acc} \\
\hline \hline
Blink Detection \cite{phuong2022eye} & 0.5837 \\
\hline
Soukupova Threshold \cite{soukupova2016eye} + Insightface   & 0.6153 \\
\hline
\textbf{OE-ConvNet (FS)} & \textbf{0.9656} \\
\hline
\textbf{LI-RI-RGB-ConvNet} & \textbf{0.9760} \\

\hline
\end{tabular}
\end{table}

Finally, 
Table~\ref{tab:Benchmark} presents a comparison between the performance obtained
by the OE-ConvNet (FS = Full Spectrum) and LI-RI-RGB-ConvNet EyeBlink detectors (our
proposals), and two existing eyeblink detectors: Blink
Detection$+$ \cite{phuong2022eye} and Soukupova Threshold~\cite{soukupova2016eye}
 $+$  InsightFace. The thresholds of both
detectors~\cite{phuong2022eye,soukupova2016eye} were retrained using the
frame-level subset of mEBAL2 and the cross-validation protocol proposed in the
mEBAL2 benchmark. The face detector of~\cite{soukupova2016eye} was updated with
the state-of-the-art detector InsightFace~\cite{deng2020retinaface}. Our
LI-RI-RGB-ConvNet proposal achieves the highest accuracy with 0.9760,
outperforming OE-ConvNet (FS). This demonstrates that our LI-RI-RGB-ConvNet
architecture enhances accuracy through late fusion~\cite{fu2021siames} of
information from the 3 cameras instead of the heterogeneous training used for
OE-ConvNet (FS). However, as a downside, this multispectral approach requires
specific hardware with three synchronized sensors. The performance of data-driven approaches such as OE-ConvNet (FS) and LI-RI-RGB-ConvNet is clearly
superior to the performance achieved by the methods Blink Detection$+$  and
Soukupova Threshold  $+$  InsightFace based on the EAR threshold (see Section \ref{sec:methods}). 

The models proposed and evaluated here are aimed to demonstrate the usefulness
of mEBAL2 for training and evaluating novel blink detectors, and in that regard compare well with recent methods like~\cite{phuong2022eye,soukupova2016eye}. Note that we are not
claiming superior performance of our models in comparison with cutting-edge
detectors based on advanced data-driven learning
architectures such as~\cite{zeng2023real,zeng2023eyelid}. 

\begin{table}[t!]
\centering
\setlength\tabcolsep{7pt}
\renewcommand{\arraystretch}{1.1}
\caption{Eyeblink detection accuracy at video-level obtained by OE-ConvLSTM and two state-of-the-art implementations \cite{phuong2022eye,soukupova2016eye}.}
\label{tab:Benchmark2}
\begin{tabular}{|l|c|}
\hline
\textbf{Method} & \textbf{Acc} \\
\hline \hline
Blink Detection+ \cite{phuong2022eye} & 0.6758 \\
\hline
Soukupova SVM \cite{soukupova2016eye} + Insightface  & 0.8145 \\
\hline
\textbf{OE-ConvLSTM} & \textbf{0.9909} \\

\hline
\end{tabular}
\end{table}

\subsubsection{mEBAL2: Blink detection at video-sequence level}

Table~\ref{tab:Benchmark2} presents the eyeblink detection accuracies at
video-sequence level of the mEBAL2 Benchmark for our OE-ConvLSTM and two
existing blink detectors: Blink Detection$+$~\cite{phuong2022eye} and Soukupova
SVM~\cite{soukupova2016eye}  $+$  InsightFace. Both
detectors~\cite{phuong2022eye,soukupova2016eye} were retrained on the
video-sequence level subset of mEBAL2,  including recalibration of Blink
Detection$+$ thresholds and parameter optimization for the Soukupova SVM model.
The results demonstrate the potential of mEBAL2 database to train eye detectors
obtaining an accuracy of 0.9909. The methods proposed
in~\cite{phuong2022eye,soukupova2016eye} improve in comparison with the
accuracies obtained at frame level. However, the performance of these methods,
based on eye landmarks, adaptive thresholds, and SVM is far from those obtained
using data-driven learning architectures such as OE-ConvLSTM.

These results are even more remarkable when considering the size of the database and the challenging e-learning environment, where pose changes are common due to the students looking at the keyboard, resulting in closed eyes appearance, as well as changes in lighting, among other factors.

\subsection{Experiments on HUST-LEBW: Evaluating the generalization ability of models trained with mEBAL2}

Table~\ref{tab:Comparison} compares different eyeblink detectors (ours vs relevant
related works) based on the public benchmark HUST-LEBW~\cite{hu2019towards}.
Our OE-ConvLSTM architecture (Proposal 4) was evaluated through two distinct
approaches: (i) OE-ConvLSTM trained using the RGB images of mEBAL2, and (ii)
OE-ConvLSTM trained using the HUST-LEBW images. The architecture trained with
mEBAL2 achieves the second-best performance in terms of the F1 metric for both
eyes, only being outperformed by the recent eyelid
method~\cite{zeng2023eyelid}. It is important to note that eyelid incorporates
a more complex structure, including an attention generator,  CNN, and LSTM
architectures (see  Section~\ref{sec:methods}). When OE-ConvLSTM is trained with
mEBAL2, a slight decrease in the performance of approximately 3\% in the F1
metric is observed, demonstrating the effectiveness of mEBAL2 for training
data-driven approaches capable of generalizing in unseen scenarios.

Our OE-ConvNet architecture (Proposals 1--3) was evaluated with different training settings on mEBAL2: Proposal $1$ was trained with both eyes using RGB and NIR images, Proposal $2$ was trained with both eyes using only RGB images, and Proposal $3$ consists of two detectors trained using RGB images for both eyes separately. Proposal $1$  outperforms the best results in the F1 score for architectures trained on mEBAL2, even surpassing our OE-ConvLSTM architecture. These results suggest the usefulness of multispectral training (RGB$+$NIR) when testing on a different dataset (note that HUST-LEBW includes RGB images only).

\begin{table}[t]
\centering
\setlength\tabcolsep{4pt}
\renewcommand{\arraystretch}{1.2}
\caption{Eyeblink detection results on the HUST-LEBW dataset~\cite{hu2019towards}. Our OE-ConvNet proposals were trained on mEBAL2 (see Table~\ref{tab:Results} for the training configuration of each Proposal). Our OE-ConvLSTM (Proposal 4) underwent two distinct training: one on mEBAL2 and the other on HUST-LEBW. The method described in \cite{soukupova2016eye} was updated using InsightFace \cite{deng2020retinaface}.}
\label{tab:Comparison}
\begin{tabular}{|c|c|c|c|c|c|}
\hline
\textbf{Training} & \textbf{Method} & \textbf{Eye} & \textbf{Recall} & \textbf{Precision} & \textbf{F1} \\
\hline \hline
\multirow{9}{*}{\rotatebox[origin=c]{90}{HUST-LEBW}} & \cite{soukupova2016eye} & Both & 0.4073 & 0.8495 & 0.5506 \\
\cline{2-6}
& \cite{phuong2022eye} & Both & 0.5899 & 0.8005 & 0.6790 \\
\cline{2-6}
& \cite{zeng2023real} & Both & 0.9764 & 0.5662 & 0.7168 \\
\cline{2-6}
& \cite{hu2019towards} & Left & 0.5410 & 0.8919 & 0.6735 \\
\cline{3-6}
& & Right & 0.4444 & 0.7671 & 0.5628 \\
\cline{2-6}
& \cite{zeng2023eyelid} & Left & 0.9180 & 0.8960 & 0.9069 \\
\cline{3-6}
& & Right & 0.9127 & 0.9274 & 0.9200 \\
\cline{2-6}
& Proposal 4 & Left & 0.8968 & 0.8014 & 0.8464 \\
\cline{3-6}
& & Right & 0.8780 & 0.7826 & 0.8276 \\
\hline
\multirow{4}{*}{\rotatebox[origin=c]{90}{mEBAL\hspace{-7mm}}} & \cite{daza2020mebal} & Left & 0.9603 & 0.6080 & 0.7446 \\[4pt]
\cline{3-6}
& & Right & 0.7950 & 0.7348 & 0.7637 \\[4pt]
\cline{3-6}
\hline
\multirow{16}{*}{\rotatebox[origin=c]{90}{mEBAL2\hspace{-40mm}}} & Proposal 1 & Left & 0.9440 & 0.7564 & 0.8399 \\
\cline{3-6}
& & Right & 0.8770 & 0.7868 & 0.8295 \\
\cline{2-6}
& Proposal 2 & Left & 0.9520 & 0.7126 & 0.8151 \\
\cline{3-6}
& & Right & 0.8934 & 0.6855 & 0.7758 \\
\cline{2-6}
& Proposal 3 & Left & 0.9200 & 0.6928 & 0.7904 \\
\cline{3-6}
& & Right & 0.9262 & 0.7152 & 0.8072 \\
\cline{2-6}
& Proposal 4 & Left & 0.8596 & 0.7656 & 0.8100 \\
\cline{3-6}
& & Right & 0.8303 & 0.7750 & 0.8017 \\
\hline
\end{tabular}
\end{table}

Proposal $2$ has the same architecture as Proposal $1$ but it
was trained only with RGB images. As a result, Proposal $2$ has lower
performance than Proposal $1$, especially for the right eye. It is
interesting because this indicates that training with more data and with both
spectra allows the creation of models with a greater generalization capacity
for different environments (different illumination, head orientation, etc.).
Furthermore, our initial approach presented in~\cite{daza2020mebal} shares the
same architecture as Proposal $2$ (OE-ConvNet). However, it was
trained with the first version of mEBAL with RGB images and therefore presents inferior
results in comparison with our Proposal $2$ trained now with mEBAL2.
Proposal $2$ improves the F1 metric in both eyes,  with 7.05\% for the
left eye and 1.21\% for the right eye. These results suggest the importance of
the usage of wide databases to train data-driven eyeblink detectors.

The training of the  OE-ConvNet architecture in Proposal $3$, which consists of two detectors trained using RGB images for both eyes separately, outperforms the performance of the other OE-ConvNet proposals in the mEBAL2 evaluation (see table~\ref{tab:Results}). However, in the HUST-LEBW evaluation, Proposal $3$ obtains worse results for F1 metrics in both eyes than the OE-ConvNet architecture trained with both eyes using RGB and NIR images (Proposal $1$). Also, for the left eye, it obtains lower F1 scores than the OE-ConvNet architecture trained with both eyes using RGB images (Proposal $2$). This result again shows the importance of training with databases with a large number of samples and how the NIR spectrum can be useful to train data-driven approaches with robust generalization capabilities.

\begin{figure}[!t]
 \centering
  \includegraphics[width=\linewidth]{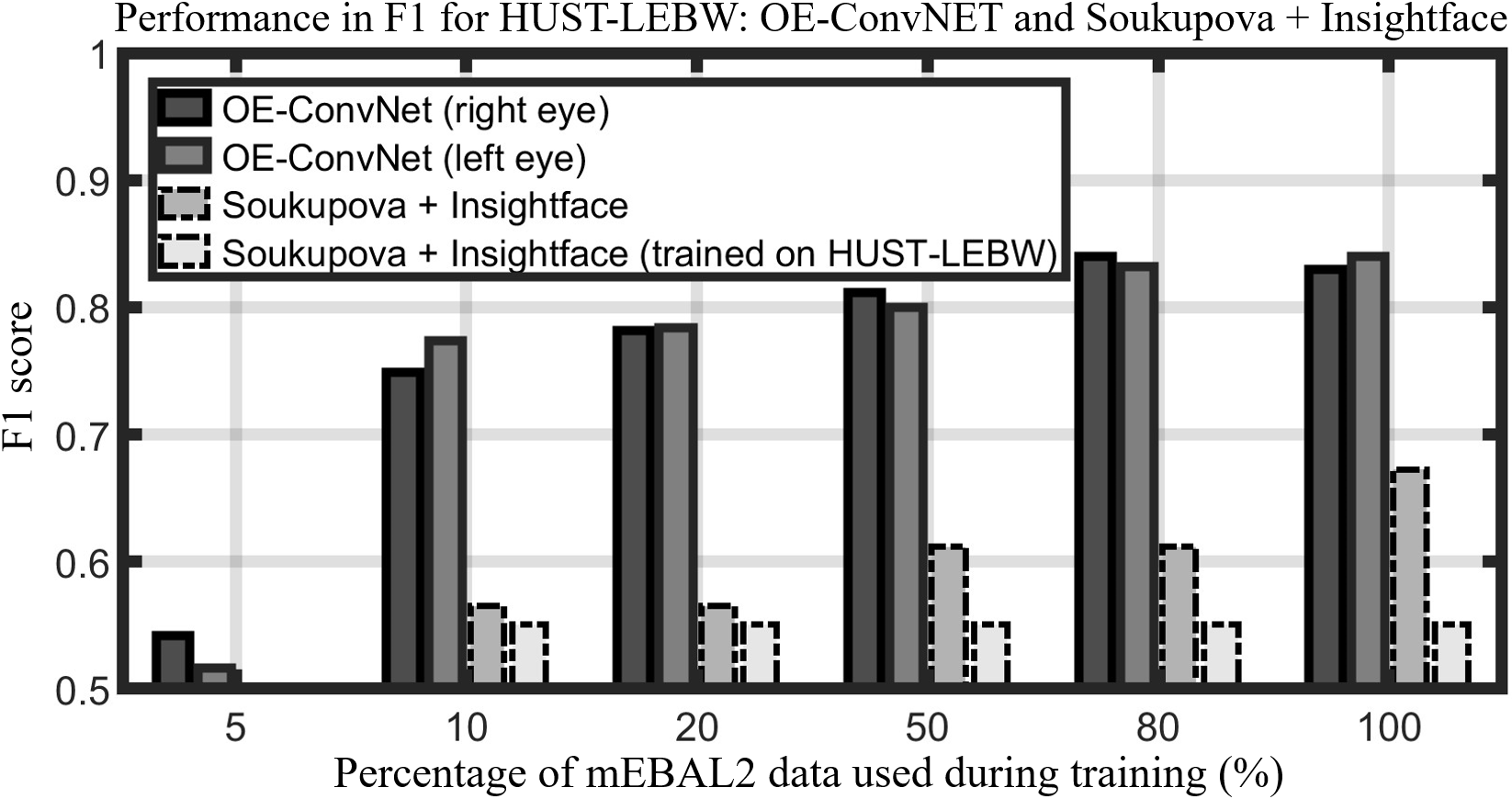} 
  \caption{F1 score results on HUST-LEBW evaluation for different training ratios in mEBAL2 for OE-ConvNet and Soukupova \cite{soukupova2016eye} + Insightface architectures.}
   \label{fig:F1}
\end{figure}

Finally, Figure~\ref{fig:F1} shows the results achieved for the F1 scores on the
HUST-LEBW dataset for different training percentages on mEBAL2 of our
OE-ConvNet method (Proposal 1) along with our adaptation of Soukupova $+$ 
InsightFace~\cite{soukupova2016eye}. The results demonstrate an increase in
accuracy for both architectures when training with a larger volume of data.
Even Soukupova $+$  InsightFace achieves better performance when trained on mEBAL2
than with HUST-LEBW training. As we can see, the large number of samples and
users in mEBAL2 allows for the improvement in the performance of eyeblink
detectors even in unconstrained scenarios.

\section{Conclusions} \label{sec:conclusions}

This work has presented a new multispectral database for
eyeblink detection. mEBAL2 is 3.52 times wider than the first version in terms
of samples and around 5 times larger in terms of users, being the largest
existing eyeblink database in the literature for research in image- and
video-based eyeblink detection and related applications, e.g.: attention level
estimation~\cite{daza2023matt} and presentation attack detection to face
biometrics~\cite{2023_Book-PAD_Face}. mEBAL2 uses visible and NIR spectra
($1$ RGB and $2$ NIR cameras). 

Besides, we explored the effects of the visible (RGB) and NIR spectra for eyeblink detection. Our results demonstrate that: (\textit{i}) the approaches trained with both spectra have a good generalization capacity for unseen scenarios, (\textit{ii}) the combination of the RGB and NIR spectrum through late fusion architectures improves the results in eyeblink detection on e-learning environments.

Our proposed architecture for blink detection at video-sequence level, based on
ConvLSTM, has achieved the highest levels of accuracy, approximately 99\%, in the
challenging e-learning environment considered in mEBAL2. Additionally, our methods achieved
the second-best performance under uncontrolled conditions in the HUST-LEBW
dataset, only surpassed by the eyelid method~\cite{zeng2023eyelid}, which is a
more complex architecture (see Section~\ref{sec:methods}).

mEBAL2 has proven to be a valuable resource to train data-driven algorithms, since a simple CNN learning architecture, when trained on mEBAL2, has demonstrated robust generalization capabilities and significantly improved results compared to its performance with mEBAL. Therefore, the results show that mEBAL2 can be used to train a new generation of data-driven approaches for eyeblink detection.
 
Future work includes: exploring in more depth the NIR spectrum, advancing in new architectures to leverage the temporal information across frames
(GRU, Transformers, etc.), exploiting modern multimodal
strategies~\cite{2023_SNCS_Human-Centric_Pena} integrating context information in the periocular region~\cite{alonsofernandez2023periocular}, exploiting recent advances in generative face biometrics ~\cite{MELZI2024102322}, and exploiting general large-scale AI models with facial analysis capabilities~\cite{ivan24chat} to provide added-value in this specific problem of eyeblink detection.

\section{Acknowledgments} \label{sec:ACKNOWLEDGMENTS}
Support by project HumanCAIC (TED2021-131787B-I00 MICINN), BIO-PROCTORING (GNOSS Program, Agreement Ministerio de Defensa-UAM-FUAM dated 29-03-2022) and Catedra ENIA UAM-VERIDAS en IA Responsable (NextGenerationEU PRTR TSI-100927-2023-2). Research partially funded by the Autonomous Community of Madrid. Roberto Daza is supported by a FPI fellowship from MINECO/FEDER. A. Morales is supported by the Madrid Government (Comunidad de Madrid-Spain) under the Multiannual Agreement with Universidad Autónoma de Madrid in the line of Excellence for the University Teaching Staff in the context of the V PRICIT (Regional Programme of Research and Technological Innovation).

 \bibliographystyle{elsarticle-num} 
 \bibliography{biography}

\begin{thebibliography}{10}
\expandafter\ifx\csname url\endcsname\relax
  \def\url#1{\texttt{#1}}\fi
\expandafter\ifx\csname urlprefix\endcsname\relax\def\urlprefix{URL }\fi
\expandafter\ifx\csname href\endcsname\relax
  \def\href#1#2{#2} \def\path#1{#1}\fi

\bibitem{2019_HBookSelfie_SuperSelfieFaceIris_Alonso}
F.~Alonso-Fernandez, R.~A. Farrugia, J.~Fierrez, J.~Bigun, Super-resolution for selfie biometrics: Introduction and application to face and iris, in: A.~Rattani, R.~Derakhshani, A.~Ross (Eds.), Selfie Biometrics, Springer, 2019, pp. 105--128.

\bibitem{alonsofernandez2023periocular}
F.~Alonso-Fernandez, J.~Bigun, J.~Fierrez, N.~Damer, H.~Proen{\c c}a, A.~Ross, Periocular biometrics: {\'A} modality for unconstrained scenarios, IEEE Comput. (2024).

\bibitem{bergasa2006real}
L.~M. Bergasa, J.~Nuevo, M.~A. Sotelo, R.~Barea, M.~E. Lopez, Real-time system for monitoring driver vigilance, IEEE Transactions on Intelligent Transportation Systems 7~(1) (2016) 63--77.

\bibitem{daza2021alebk}
R.~Daza, D.~DeAlcala, A.~Morales, R.~Tolosana, R.~Cobos, J.~Fierrez, {ALEBk: Feasibility study of attention level estimation via blink detection applied to e-learning}, in: Proc. AAAI Workshop on Artificial Intelligence for Education, 2022.

\bibitem{daza2023matt}
R.~Daza, L.~F. Gomez, A.~Morales, J.~Fierrez, R.~Tolosana, R.~Cobos, J.~Ortega-Garcia, {MATT: Multimodal Attention Level Estimation for e-learning Platforms}, in: Proc. AAAI Workshop on Artificial Intelligence for Education, 2023.

\bibitem{rosenfield2011computer}
M.~Rosenfield, Computer vision syndrome: a review of ocular causes and potential treatments, Ophthalmic and Physiological Optics 31~(5) (2011) 502--515.

\bibitem{jung2020deepvision}
T.~Jung, S.~Kim, K.~Kim, Deepvision: Deepfakes detection using human eye blinking pattern, IEEE Access (2020) 83144--83154.

\bibitem{hernandez2019edbb}
J.~Hernandez-Ortega, R.~Daza, A.~Morales, J.~Fierrez, J.~Ortega-Garcia, {edBB: Biometrics and Behavior for Assessing Remote Education}, in: Proc. AAAI Workshop on Artificial Intelligence for Education, 2020.

\bibitem{daza2023edbb}
R.~Daza, A.~Morales, R.~Tolosana, L.~F. Gomez, J.~Fierrez, J.~Ortega-Garcia, {edBB-Demo: Biometrics and Behavior Analysis for Online Educational Platforms}, in: Proc. AAAI Conf. on Artificial Intelligence (Demonstration), 2023.

\bibitem{becerra2023m2lads}
{\'A}.~Becerra, R.~Daza, R.~Cobos, A.~Morales, M.~Cukurova, J.~Fierrez, {M2LADS: A System for Generating MultiModal Learning Analytics Dashboards in Open Education}, in: Proc. Annual Computers, Software, and Applications Conference (COMPSAC) in the Workshop on Open Education Resources, 2023.

\bibitem{bagley1979effect}
J.~Bagley, L.~Manelis, Effect of awareness on an indicator of cognitive load, Perceptual and Motor Skills 49~(2) (1979) 591--594.

\bibitem{holland1972blinking}
M.~K.~Holland, G.~Tarlow, Blinking and mental load, Psychological Reports 31~(1) (1972) 119--127.

\bibitem{daza2020mebal}
R.~Daza, A.~Morales, J.~Fierrez, R.~Tolosana, {mEBAL: A Multimodal Database for Eye Blink Detection and Attention Level Estimation}, in: International Conference on Multimodal Interaction, 2020, pp. 32--36.

\bibitem{hu2019towards}
G.~Hu, Y.~Xiao, Z.~Cao, L.~Meng, Z.~Fang, J.~T. Zhou, J.~Yuan, Towards real-time eyeblink detection in the wild: Dataset, theory and practices, IEEE Transactions on Information Forensics and Security (2019) 2194--2208.

\bibitem{zeng2023real}
W.~Zeng, Y.~Xiao, S.~Wei, J.~Gan, X.~Zhang, Z.~Cao, Z.~Fang, J.~T. Zhou, Real-time multi-person eyeblink detection in the wild for untrimmed video, in: Proceedings of the IEEE/CVF Conference on Computer Vision and Pattern Recognition, 2023, pp. 13854--13863.

\bibitem{2022_INFFUS_Periocular_Alonso}
F.~Alonso-Fernandez, K.~Raja, R.~Raghavendra, C.~Busch, J.~Bigun, R.~Vera-Rodriguez, J.~Fierrez, Cross-sensor periocular biometrics for partial face recognition in a global pandemic: Comparative benchmark and novel multialgorithmic approach, Information Fusion 83-84 (2022) 110--130.

\bibitem{Talkingface}
{Talking Face}, Talking face, \url{https://personalpages.manchester.ac.uk/staff/timothy.f.cootes/data/talking_face/talking_face.html}, accessed: 2023-05-25 (2021).

\bibitem{pan2007eyeblink}
G.~Pan, L.~Sun, Z.~Wu, S.~Lao, {Eyeblink-based Anti-spoofing in Face Recognition from a Generic Webcamera}, in: Proc. IEEE International Conference on Computer Vision, 2007.

\bibitem{drutarovsky2014eye}
T.~Drutarovsky, A.~Fogelton, {Eye Blink Detection using Variance of Motion Vectors}, in: Proc. European Conference on Computer Vision, 2014, pp. 436--448.

\bibitem{radlak2015silesian}
K.~Radlak, M.~Bozek, B.~Smolka, {Silesian Deception Database: Presentation and Analysis}, in: Proc. ACM on Workshop on Multimodal Deception Detection, 2015, pp. 29--35.

\bibitem{anas2017online}
E.~R. Anas, P.~Henriquez, B.~J. Matuszewski, {Online Eye Status Detection in the Wild with Convolutional Neural Networks}, in: Proc. International Conf. on Computer Vision Theory and Applications, 2017, pp. 88--95.

\bibitem{phuong2022eye}
T.~T. Phuong, L.~T. Hien, N.~D. Vinh, {An Eye Blink Detection Technique in Video Surveillance based on Eye Aspect Ratio}, in: Proc. Advanced Communication Technology, 2022, pp. 534--538.

\bibitem{soukupova2016eye}
T.~Soukupovà, J.~Cech, {Eye Blink Detection using Facial Landmarks}, in: Proc. Computer Vision Winter Workshop, 2016.

\bibitem{carion2020end}
N.~Carion, F.~Massa, G.~Synnaeve, N.~Usunier, A.~Kirillov, S.~Zagoruyko, {End-to-end Object Detection with Transformers}, in: Proc. European Conference on Computer Vision (ECCV), 2020.

\bibitem{zeng2023eyelid}
W.~Zeng, Y.~Xiao, G.~Hu, Z.~Cao, S.~Wei, Z.~Fang, J.~T. Zhou, J.~Yuan, Eyelid’s intrinsic motion-aware feature learning for real-time eyeblink detection in the wild, Transactions on Information forensics and security 18 (2023) 5109--5121.

\bibitem{schiffman1990sensation}
H.~Richard~Schiffman (Ed.), Sensation and Perception: An Integrated Approach, John Wiley \& Sons, 1990.

\bibitem{kirschstein2009source}
T.~Kirschstein, R.~K{\"o}hling, What is the source of the {EEG}?, Clinical EEG and Neuroscience 40~(3) (2009) 146--149.

\bibitem{hall2020guyton}
J.~E. Hall, M.~E. Hall (Eds.), Guyton and Hall Textbook of Medical Physiology e-Book, Elsevier Health Sciences, 2020.

\bibitem{govindasamy2001successful}
T.~Govindasamy, Successful implementation of e-learning: Pedagogical considerations, The Internet and Higher Education 4~(3-4) (2001) 287--299.

\bibitem{deng2020retinaface}
J.~Deng, et.al, {{RetinaFace}: Single-Shot Multi-Level Face Localisation in the Wild}, in: Proc. IEEE/CVF Conf. on Computer Vision and Pattern Recognition, 2020, pp. 5203--5212.

\bibitem{dong2018supervision}
X.~Dong, S.-I. Yu, X.~Weng, S.-E. Wei, Y.~Yang, Y.~Sheikh, {Supervision-by-Registration: An Unsupervised Approach to Improve the Precision of Facial Landmark Detectors}, in: Proc. IEEE Conference on Computer Vision and Pattern Recognition, 2018, pp. 360--368.

\bibitem{simonyan2014very}
K.~Simonyan, A.~Zisserman, Very deep convolutional networks for large-scale image recognition, arXiv preprint (2014).

\bibitem{wei2016convolutional}
S.-E. Wei, V.~Ramakrishna, T.~Kanade, Y.~Sheikh, {Convolutional Pose Machines}, in: Proc. IEEE Conference on Computer Vision and Pattern Recognition, 2016, pp. 4724--4732.

\bibitem{dlib}
{Dlib}, Dlib c++ library, \url{http://dlib.net/}, accessed: 2024-02-20 (2024).

\bibitem{2015_FSI_FacialSoftBio_Pedro}
P.~Tome, R.~Vera-Rodriguez, J.~Fierrez, J.~Ortega-Garcia, Facial soft biometric features for forensic face recognition, Forensic Science International 257 (2015) 171--284.

\bibitem{shi2015convolutional}
X.~Shi, Z.~Chen, H.~Wang, D.-Y. Yeung, W.-K. Wong, W.-c. Woo, Convolutional lstm network: A machine learning approach for precipitation nowcasting, Advances in Neural Information Processing Systems 28 (2015).

\bibitem{2018_INFFUS_MCSreview1_Fierrez}
J.~Fierrez, A.~Morales, R.~Vera-Rodriguez, D.~Camacho, Multiple classifiers in biometrics. part 1: Fundamentals and review, Information Fusion 44 (2018) 57--64.

\bibitem{fu2021siames}
K.~Fu, D.-P. Fan, G.-P. Ji, Q.~Zhao, J.~Shen, C.~Zhu, Siamese network for rgb-d salient object detection and beyond, Transactions on pattern analysis and machine intelligence 44~(9) (2021) 5541--5559.

\bibitem{2023_Book-PAD_Face}
J.~Hernandez-Ortega, J.~Fierrez, A.~Morales, J.~Galbally, Introduction to presentation attack detection in face biometrics and recent advances, in: S.~Marcel, J.~Fierrez, N.~Evans (Eds.), Handbook of Biometric Anti-Spoofing, Springer, 2023, 3rd Ed.

\bibitem{2023_SNCS_Human-Centric_Pena}
A.~Peña, I.~Serna, A.~Morales, J.~Fierrez, A.~Ortega, A.~Herrarte, M.~Alcantara, J.~Ortega-Garcia, Human-centric multimodal machine learning: Recent advances and testbed on ai-based recruitment, SN Computer Science 4~(5) (2023) 434.

\bibitem{MELZI2024102322}
P.~Melzi, R.~Tolosana, R.~Vera-Rodriguez, M.~Kim, C.~Rathgeb, X.~Liu, I.~DeAndres-Tame, A.~Morales, J.~Fierrez, et~al., {FRCSyn-onGoing}: Benchmarking and comprehensive evaluation of real and synthetic data to improve face recognition systems, Inf. Fusion 107 (2024) 102322.

\bibitem{ivan24chat}
I.~Deandres-Tame, R.~Tolosana, R.~Vera-Rodriguez, A.~Morales, J.~Fierrez, J.~Ortega-Garcia, How good is {ChatGPT} at face biometrics? a first look into recognition, soft biometrics, and explainability, IEEE Access 12 (2024) 34390--34401.

\end{thebibliography}





\end{document}